\documentclass[twocolumn]{article}

\usepackage[utf8]{inputenc}
\usepackage[
backend=biber,
style=alphabetic,
sorting=ynt
]{biblatex}
\usepackage{enumitem}

\usepackage{hyperref}
\usepackage{xurl}
\hypersetup{
    breaklinks=true,   
    colorlinks=true,   
    linkcolor=blue,
    filecolor=magenta,      
    urlcolor=cyan,
    pdfpagemode=FullScreen,
}

\title{Semantic Table Detection with LayoutLMv3}
\author{Ivan Silajev \\ \href{mailto@ivan.silajev@warwick.ac.uk}{ivan.silajev@warwick.ac.uk}
        \and Niels Victor \\ \href{mailto@niels@accelextech.com}{niels@accelextech.com}
        \and Phillip Mortimer \\ \href{mailto@phillip@accelextech.com}{phillip@accelextech.com}
        }
\date{October 2022}

\begin{document}

\twocolumn[
    \begin{@twocolumnfalse}
    \maketitle
    \begin{abstract}
\noindent
This paper presents an application of the LayoutLMv3 model for semantic table detection on financial documents from the IIIT-AR-13K dataset.
The motivation behind this paper's experiment was that LayoutLMv3's official paper had no results for table detection using semantic information.
We concluded that our approach did not improve the model's table detection capabilities, for which we can give several possible reasons. Either the model's weights were unsuitable for our purpose, or we needed to invest more time in optimising the model's hyperparameters. It is also possible that semantic information does not improve a model's table detection accuracy.
    \end{abstract}
    \vspace{10pt}
    \end{@twocolumnfalse}
]

\section{Introduction}
Over the last few years, the appearance of powerful new models, such as BERT \cite{https://doi.org/10.48550/arxiv.1810.04805} and GPT-3 \cite{https://doi.org/10.48550/arxiv.2005.14165}, significantly advanced the state-of-the-art (SOTA) frontier of the natural language processing (NLP) field.
Most of these models process text as a one-dimensional array, but most real-world documents require the 2D location of text on the page and visual information such as font style, size and colour to read the document correctly.
Automating the reading of these Visually Rich Documents (VRDs) requires multi-modal neural networks that can infer from semantic and image data.
VRD specialised multi-modal neural network models is a field of active research that includes recent advances such as LAMBERT \cite{Garncarek_2021}, TILT \cite{https://doi.org/10.48550/arxiv.2102.09550} and LayoutLMv3 \cite{https://doi.org/10.48550/arxiv.2204.08387}.\\

An essential task in financial VRD understanding is table detection (TD).
The most successful models in this field include object-detection models such as Faster R-CNN \cite{https://doi.org/10.48550/arxiv.1506.01497} and DiT \cite{https://doi.org/10.48550/arxiv.2203.02378}, which only use image and spatial features to determine table locations.
The creators of LayoutLMv3 also tested their model for TD and other object detection tasks, but they used no semantic data in those instances, and the dataset they used, PubLayNet \cite{https://doi.org/10.48550/arxiv.1908.07836}, was not necessarily composed of financial documents.
To our knowledge, no previous work successfully used semantic features to improve a model's TD performance.
Intuitively these features should be helpful for the task; for example, figure captions and subtotal row headers could provide a signal to help a model detect a table's location.\\

We \textbf{hypothesised} that using this semantic information could improve a model's TD capability.\\

This paper documents how we tested our hypothesis using the LayoutLMv3 model and IIIT-AR-13K, a predominantly financial document dataset, and what we concluded from our results.
Aside from the potential task improvement, using multi-modal VRD models for semantic TD can enable future applications of performing TD in parallel with other text-processing tasks, such as key information extraction.
Despite coming to an answer that disproves our hypothesis, we believe that this paper can help anyone who wants to design new experiments involving semantic TD.\\

The benefits of this paper include the following:
\begin{itemize}[noitemsep,topsep=0pt,parsep=0pt,partopsep=0pt]
    \item The application of LayoutLMv3 to IIIT-AR-13K
    \item Performance metrics from a SOTA VRD model on semantic TD
    \item Details of an experiment for carrying out semantic TD
\end{itemize}

\section{Related Work} \label{Related Work}
The authors of LayoutLMv3 published only one pre-trained version of the model specialised for object detection, trained on the PubLayNet dataset but not exposed to semantic information.
We predicted that introducing this pre-trained model to a new modality could significantly change the trained weights.
For example, such changes may include the loss of visual information carried through the model, as the training procedure would inject some tensors with semantic information.
We did not find other pre-trained versions of LayoutLMv3 that had a mixture of image and text modality for semantic object detection.\\

While working on our experiment, another group of researchers published a paper on a new table detection model called RobusTabNet \cite{Ma_2023}.
The paper introduces an architecture that combines a table detector and table structure recogniser for efficiently detecting tables in any document type.
The table detector comprises a CornerNet \cite{https://doi.org/10.48550/arxiv.1808.01244} architecture sandwiched between a ResNet-18 \cite{https://doi.org/10.48550/arxiv.1512.03385} backbone and an R-CNN.
In addition, the table structure recogniser solidifies the model's guess of a table's bounding box.
Despite the model being image-only, we propose, in this paper, how one can integrate semantic features into the model as well.

\section{Our Approach}

Before proceeding, we clarify the hypothesis we want to test: Does semantic information on top of image and layout information improve a transformer-based model's performance in TD for financial documents?
To test it, we required a multi-modal model that combined the language modelling capabilities of modern BERT variants and the layout and image modelling capabilities of DiT.
LayoutLMv3 was the newest version of transformer models of its kind that satisfied our requirements, justifying our use of it.
We used the IIIT-AR-13K dataset for our experiment, as it is specialised for object detection tasks in financial documents, including table detection.

\begin{table*}[t]
    \begin{center}
        \begin{tabular}{c|c|c|c|c}
            \textbf{Model} & \textbf{Max} & \textbf{Gradient} & \textbf{Base} & \textbf{Weight}\\
            \textbf{Type} & \textbf{Training} & \textbf{Accumulation} & \textbf{Learning} & \textbf{Decay}\\
            & \textbf{Iterations} & \textbf{Steps} & \textbf{Rate} & \textbf{Rate}\\
            \hline
            LMv3 (Image-only) & 100000 & 8 & 1.30e-05 & 0.05\\
            LMv3 (With text) & 100000 & 8 & 3.68e-05 & 0.10
        \end{tabular}
        \caption{\label{tab:table2} Hyperparameters of image-only and text-processing versions of LayoutLMv3 optimised for TD.}
    \end{center}
    \vspace{-5mm}
\end{table*}

\subsection{Procedure Details}
To specialise LayoutLMv3 for object detection, we used it within a Detectron2 \cite{wu2019detectron2} framework with the same configurations and model weights as the pre-trained LayoutLMv3 model trained on the PubLayNet dataset.
We made no architectural changes to the model and modified only the hyperparameters that affected the model training procedure.
We presumed that a sufficiently high weight decay would relax the model's weights enough to learn new features from the text data, so we varied it for our tests.
With such a large model, we supported the retraining with more training iterations.\\

We first trained the model on the IIIT-AR-13K dataset without using any text information and then with text.
As our performance measure, we used the average precision (AP), defined by the COCO \cite{coco} standard, with which the model predicts the bounding box of the tables.
Additionally, we optimised the hyperparameters for both models, as it made little sense to compare their results if neither were at their peak performance.
Specifically, we tested the effects of varying the maximum number of training iterations, the number of gradient accumulation steps, the base learning rate and the weight decay rate.

\subsection{Experiment}
As predicted in Section \ref{Related Work}, the model's predictive capabilities changed upon exposing it to semantic information, likely indicating that we successfully modified its modality. But we did not expect that the model would yield lower AP metrics after the retraining, even after optimising its hyperparameters, as seen in Table \ref{tab:table1}.
In Table \ref{tab:table2}, we compare the AP metrics between both versions of LayoutLMv3 that we tested and RobusTabNet's AP metrics on the validation and test IIIT-AR-13K datasets.

\begin{table}[ht!]
    \begin{center}
        \begin{tabular}{c|c|c}
            \textbf{Model Type} & \textbf{bbox AP} & \textbf{bbox AP}\\
            & \textbf{Validation} & \textbf{Test}\\
            \hline
            LMv3 (Image-only) & 92.2 & 92.9\\
            LMv3 (With text) & 92.1 & 92.9\\
            \textbf{RobusTabNet} & \textbf{98.2} & \textbf{97.7}
        \end{tabular}
        \caption{\label{tab:table1} Max Average Precision of LayoutLMv3 for TD task on IIIT-AR-13K depending on modality. RobusTabNet is shown for SOTA.}
    \end{center}
    \vspace{-10mm}
\end{table}

\section{Evaluation \& Conclusion}
We concluded that retraining the LayoutLMv3 model with semantic information did not improve its table detection accuracy with our methodology. We propose three possible reasons for this result.\\

Firstly, the fact that LayoutLMv3's paper \cite{https://doi.org/10.48550/arxiv.2204.08387} did not give any results for semantic TD and that we did not improve on its results when using semantics could suggest that semantic information does not help in TD.
We thought textual elements in row and column headers could strongly indicate a table's presence.
However, it could be that, naturally, TD is an intensely visual task, in which case we would not expect semantic information to improve a model's performance.
The much stronger results of RobusTabNet, as seen in Table \ref{tab:table1}, reinforce this possibility, as it only does visual modelling in contrast to LayoutLMv3's additional language modelling capabilities.\\
The second possibility is that we may not have invested enough time and resources to overcome the lack of pre-trained semantic table detectors.
Current trends in pre-training tend towards long training runs on massive datasets.
We could not run such long-term, large-scale runs with the time frame and resources allocated to this project.
Furthermore, the available financial table detection datasets are limited, with FinTabNet \cite{zheng2020global} and IIIT-AR-13K being the two main ones available.
Combined, they are much smaller than some other general table detection datasets.\\

The third possibility, following closely from the second, is that we may not have invested enough time in optimising the model's hyperparameters.
With more testing, we could have found a combination of hyperparameters that would have adequately relaxed the pre-trained weights of the model.
Such a combination could have let it learn the semantic modality for the TD task more efficiently.

\subsection{Future Directions}
The three reasons listed for our result provide starting points for future research. The insignificance of semantic information for table detection would suggest that TD is a visually-dominated task and that vision-based models such as RobusTabNet are a more promising future research direction.
Alternatively, one could work more to expand the universe of pre-trained models and financial table detection datasets.
Such expansions would have potential benefits for tasks beyond just semantic TD.
Lastly, one can carry out a more thorough hyperparameter test using Table \ref{tab:table1} as a starting point.\\

Another promising avenue lies in modifying the RobusTabNet model by integrating it with more modern architectures, for example, by replacing RobusTabNet's ResNet-18 backbone with LayoutLMv3 in the table detector and table structure recogniser.

\printbibliography

\end{document}